\documentclass[letterpaper]{article} 
\usepackage{aaai2026}  
\usepackage{times}  
\usepackage{helvet}  
\usepackage{courier}  
\usepackage[hyphens]{url}  
\usepackage{graphicx} 
\usepackage{natbib}  
\usepackage{caption} 
\usepackage{algorithm}
\usepackage{algorithmic}
\usepackage{amsfonts}
\usepackage{amsmath}
\usepackage{multirow}
\usepackage{newfloat}
\usepackage{listings}
\DeclareCaptionStyle{ruled}{labelfont=normalfont,labelsep=colon,strut=off} 
\floatstyle{ruled}
\newfloat{listing}{tb}{lst}{}
\floatname{listing}{Listing}
\nocopyright

\title{Faster and Better: Reinforced Collaborative Distillation and\\ Self-Learning for Infrared-Visible Image Fusion}

\author {
	Yuhao Wang,
	Lingjuan Miao,
	Zhiqiang Zhou\thanks{Corresponding author},
	Yajun Qiao,
	Lei Zhang
}
\affiliations {
	School of Automation, Beijing Institute of Technology, Beijing, China\\
	wayhao@bit.edu.cn, miaolingjuan@bit.edu.cn, zhzhzhou@bit.edu.cn,
	yajun@bit.edu.cn, zhangleiii@bit.edu.cn
}

\usepackage{bibentry}

\begin{document}

\maketitle

\begin{abstract}
Infrared and visible image fusion plays a critical role in enhancing scene perception by combining complementary information from different modalities. Despite recent advances, achieving high-quality image fusion with lightweight models remains a significant challenge. To bridge this gap, we propose a novel collaborative distillation and self-learning framework for image fusion driven by reinforcement learning. Unlike conventional distillation, this approach not only enables the student model to absorb image fusion knowledge from the teacher model, but more importantly, allows the student to perform self-learning on more challenging samples to enhance its capabilities. Particularly, in our framework, a reinforcement learning agent explores and identifies a more suitable training strategy for the student. The agent takes both the student's performance and the teacher-student gap as inputs, which leads to the generation of challenging samples to facilitate the student's self-learning. Simultaneously, it dynamically adjusts the teacher’s guidance strength based on the student's state to optimize the knowledge transfer. Experimental results demonstrate that our method can significantly improve student performance and achieve better fusion results compared to existing techniques.
\end{abstract}


\section{Introduction}
Infrared and visible image fusion (IVIF) aims to integrate complementary information from different spectral modalities to enhance scene perception, which has been employed in various real-world applications, including robotics, autonomous driving, and remote sensing \cite{liu2024infrared}. 

Due to the powerful feature extraction capabilities of neural networks, deep learning-based fusion techniques have become dominant. Although significant progress has been made with current methods, achieving superior fusion quality with lightweight models remains a major challenge. This limitation severely restricts their applicability, particularly in real-time or resource-constrained scenarios.

A promising solution to this issue is knowledge distillation, which enables lightweight student models to learn from well-trained teacher models. However, existing methods typically adopt a teacher-dominated paradigm, where the student model passively mimics the teacher's outputs without opportunities for independent learning. This limitation motivates us to explore a collaborative framework that integrates both knowledge distillation and self-learning for the student model.

\begin{figure}[t]
	\centering
	\includegraphics[width=8.5cm]{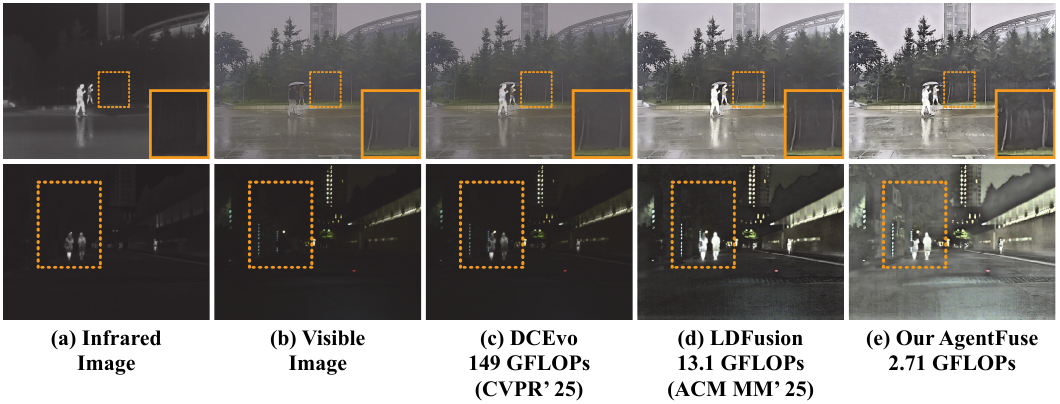} 
	\caption{Some examples of fusion results obtained by our AgentFuse and other state-of-the-art fusion methods. From (a) to (e): infrared image, visible image, and fused images from DCEVo \cite{liu2025dcevo}, LDFusion (teacher) \cite{wang2024infrared}, and our AgentFuse.}
	\label{fig:head}
\end{figure}

As we know, human learning benefits from both teacher guidance and independent study. We can acquire essential knowledge from teachers, but due to inevitable gaps between individuals, we may not fully absorb everything taught. However, through self-learning, we can reinforce this learned knowledge and, in some cases, even surpass our teachers by engaging in more challenging exercises. 

We find that this principle can also be applied to the training of student networks. By coordinating knowledge distillation and self-learning, student models with significantly fewer parameters than their teachers can still achieve, or even exceed, the performance of the teacher model (see the examples in Fig. \ref{fig:head}).

To facilitate this collaborative process, we integrate a reinforcement learning (RL) agent into the framework. RL agents can discover optimal policies based on carefully designed reward functions, making them well-suited for identifying effective training strategies for the student model. More specifically, the agent adaptively generates more challenging training samples to promote self-learning, enabling the student to extract more robust features and develop effective fusion strategies. This enhances the student's generalizability and adaptiveness to challenging scenarios, such as low-light conditions.

During the self-learning process, the agent also adjusts the strength of the teacher’s guidance to perform adaptive knowledge transfer. This dynamic interaction allows the student to benefit from both teacher direction and independent exploration, leading to a more effective unlocking of its potential. The main contributions are summarized as follows: 
\begin{itemize}

	\item{We propose a reinforced collaborative distillation and self-learning framework for IVIF, which enables the student model to self-learn with appropriately challenging samples. This improves feature extraction and enhances the model's ability to handle complex scenarios, thereby unlocking its maximum performance potential. }
	
	\item {We integrate a RL agent into the framework to enable adaptive knowledge transfer and enhance student self-learning. To the best of our knowledge, this is the first work to leverage reinforcement learning for model lightweighting in IVIF, revealing its potential in advancing IVIF performance.}
	
	\item{Extensive experiments demonstrate that our approach significantly improves both fusion performance and inference speed, highlighting the strong application potential of the proposed fusion model and framework.}
\end{itemize}

\begin{figure*}[t]
	\centering
	\includegraphics[width=12cm]{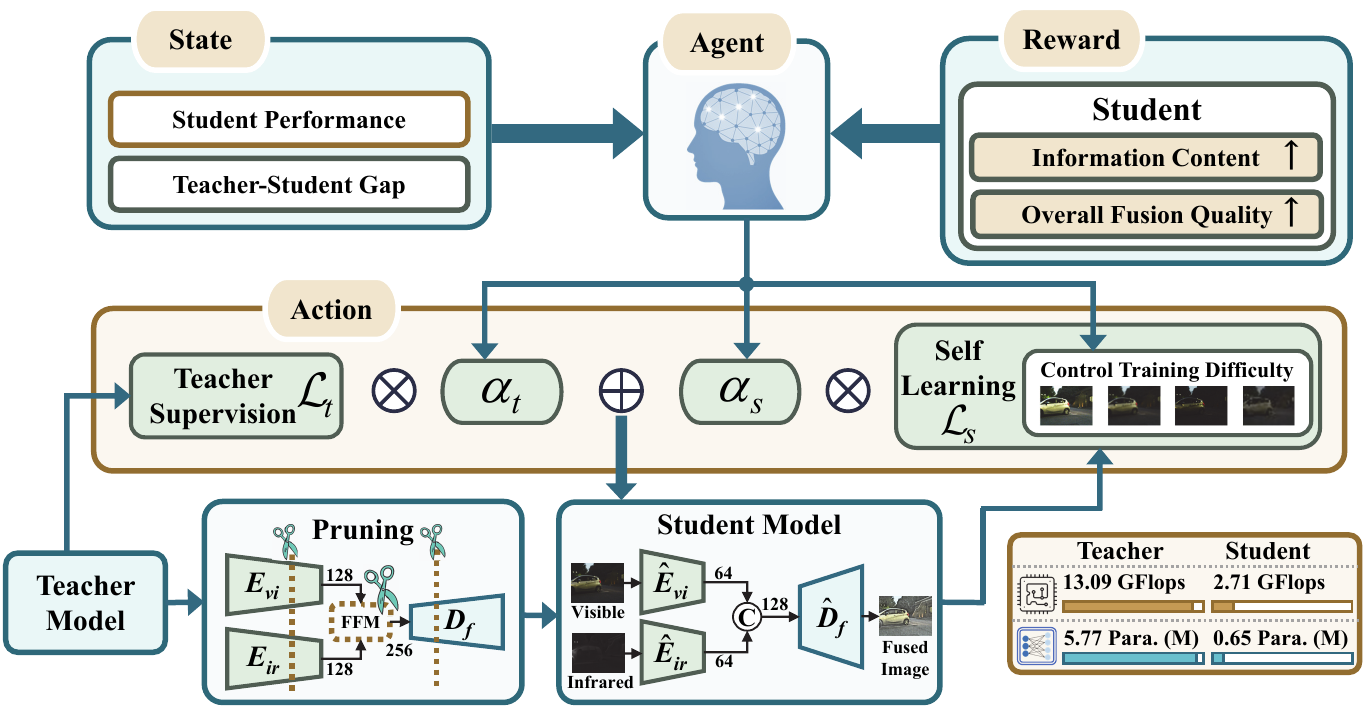} 
	\caption{The framework of the proposed method.  The student model processes visible and infrared images through separate encoding layers ($\hat E_{vi}$ and $\hat E_{ir}$), followed by a feature concatenation. The fused features are then decoded by $\hat D_{f}$ to produce the final fused image.}
	\label{fig:overall}
\end{figure*}

\section{Related Works}
\subsection{Deep Learning-Based IVIF Methods.} 
Infrared and visible image fusion (IVIF) plays a pivotal role in enhancing scene perception by integrating complementary source information. In recent years, deep learning-based methods have become the dominant approach to addressing the inherent challenges of IVIF. Early fusion approaches \cite{li2018densefuse} tend to utilize autoencoders to independently fuse and reconstruct the source image features. However, these methods are typically not end-to-end, which limits the optimization of the fusion model and consequently restricts the overall fusion performance. To address these limitations, in \cite{xu2020u2fusion}, a end-to-end convolutional neural network (CNN)-based general image fusion method was proposed. Recently proposed approaches often use more sophisticated networks, such as transformers \cite{chang2023aft, yi2024text} or hybrid transformer-CNN models \cite{chen2025sdsfusion, zhao2023cddfuse}, to enhance fusion performance. However, these methods rely heavily on various manual designs, making them rather complex, and usually can not achieve satisfactory fused images in the cases that were not considered. Despite these promising developments, achieving high-quality image fusion with lightweight models remains a significant challenge in the field.

\subsection{Knowledge Distillation} 
Knowledge distillation (KD) has emerged as a promising paradigm for model optimization, wherein a compact student model is trained to mimic the behavior of a larger, more accurate teacher network. Existing KD approaches for IVIF can be broadly categorized into pixel-wise and feature-based distillation methods. Pixel-wise knowledge distillation \cite{xue2024novel} directly enforces the student to replicate the teacher’s output at the pixel level. While this strategy facilitates precise output matching, it often fails to transfer the teacher's rich semantic knowledge, thus limiting the student's representational capacity. In contrast, feature-based knowledge distillation (\textit{e.g.}, \cite{yang2025kdfuse, wu2025every}) requires human selection of intermediate features as distillation targets, a process that is typically guided by empirical intuition or trial-and-error, and often necessitates the design of complex distillation strategies. However, most current KD methods \cite{xiang2025dkdm,zhou2025dynamic} focus on aligning the student’s behavior with that of the teacher, overlooking the potential for the student to learn autonomously. This excessive reliance on teacher supervision constrains the student’s capacity to learn more effective feature representations, ultimately limiting overall performance.

\begin{figure*}[t]
	\centering
	\includegraphics[width=12cm]{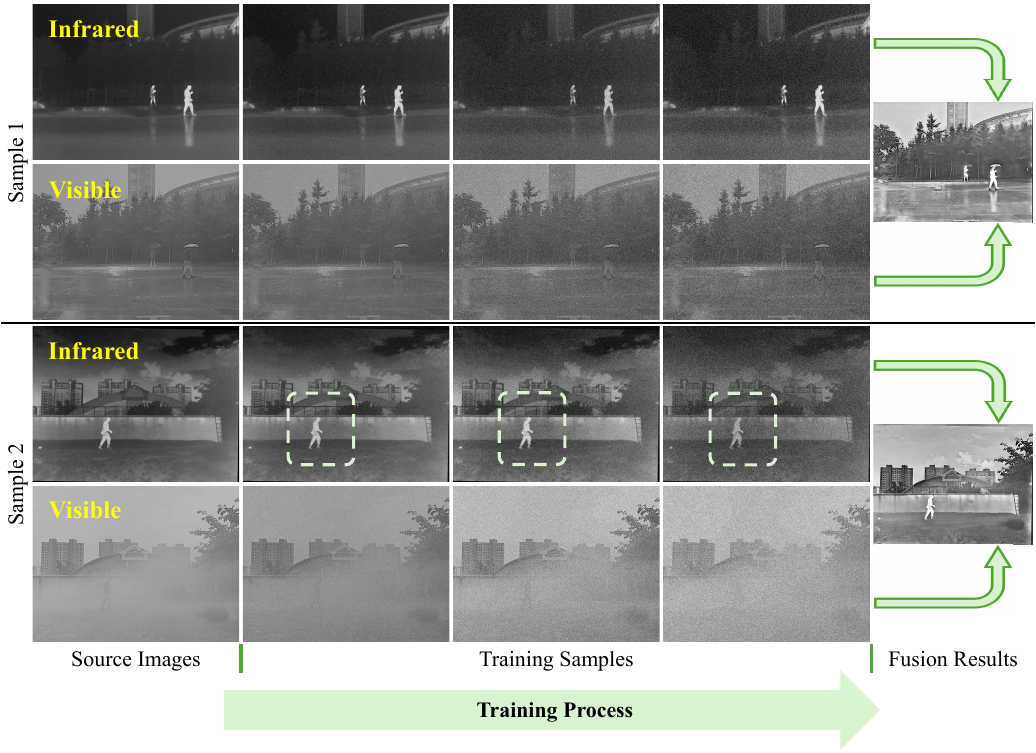} 
	\caption{Some examples of generated challenging training samples.}
	\label{fig:sample}
\end{figure*}

\section{Methodology}
\subsection{Overall Framework}
Fig. \ref{fig:overall} illustrates the overall framework of our method, where the reinforcement learning agent acts  as the central coordinator, orchestrating teacher guidance while simultaneously promoting autonomous learning in the student.

\textbf{Teacher Guidance.} 
To facilitate effective knowledge transfer, we encourage the student's output to align with the teacher's at the feature level, rather than only at the final output. This feature-level alignment enables the student model to capture richer and more detailed intermediate representations from the teacher, thereby narrowing the gap between their fusion capabilities. Specifically, we utilize a pre-trained VGG network \cite{simonyan2014very} to extract feature maps from both the teacher’s and the student’s fused images, and minimize the feature discrepancy through the teacher-guidance loss:

\begin{equation}
	\label{eq:lt}
	\mathcal{L}_t = \sum\limits_m {\left( {{{\left\| {\mathbf{f} _m^s - \mathbf{f} _m^t} \right\|}_2}} \right)}, 
\end{equation}
where $\mathbf{f} _m^s$ and $\mathbf{f} _m^t$ denote the feature maps of student’s fused image and teacher’s fused image, respectively, extracted by the \textit{m}-th convolution layer of the pre-trained VGG-19 network.

\textbf{Student Self-Learning.} 
Unlike existing KD methods \cite{yang2025kdfuse,wu2025every,yang2025multi}, our approach places greater emphasis on the student model's ability to perform self-learning on more challenging samples. By engaging with these more difficult training data, the student model is encouraged to learn more robust and discriminative feature representations, which not only improves generalization performance but also enhances its adaptability to complex real-world scenarios.

To achieve this, we intentionally degrade the quality of the source images using various techniques, such as Gaussian blurring, brightness adjustment, and lossy compression. The level of degradation is dynamically adjusted by a RL agent based on the student's current learning status (will be detailed in next subsection). The student is then encouraged to generate consistent fusion outputs for both the original and degraded inputs. This self-supervised consistency constraint enables the student to develop more robust and effective fusion strategies, rather than merely imitating the teacher, which can be achieved through a self-learning loss defined as follows:

\begin{equation}
	\label{eq:ls}
	\mathcal{L}_s = \frac{1}{N}{\sum\limits_{i = 1}^N {\left( {\hat{\mathbf{I}}_s - {\mathbf{I}_s}} \right)} ^2},
\end{equation}
where $\hat{\mathbf{I}}_s$ and $\mathbf{I}_s$ denote the student’s fusion results on the original and degraded inputs, respectively, and $N$ is the number of pixels. 

\textbf{Overall Loss.}
The total loss function combines the teacher-guidance loss and the self-learning loss:
\begin{equation}
	\label{eq:la}
	\mathcal{L}_a = \alpha_t \cdot \mathcal{L}_t + \alpha_s \cdot \mathcal{L}_s,
\end{equation}
where $\alpha_t$ and $\alpha_s$ are weighting coefficients dynamically adjusted by an agent, as detailed in the following subsection. This formulation allows the student model to both benefit from teacher guidance and explore on its own, thereby improving its fusion performance.

The student network adopts a compact dual-branch architecture, obtained by pruning the teacher network (LDFusion \cite{wang2024infrared}). In this way, the student model not only inherits prior knowledge from the teacher but also achieves substantial parameter reduction. Through this pruning process, up to $88\%$ of the parameters are removed. Remarkably, despite this significant reduction in model size, the student network still achieves superior fusion performance compared to the teacher (see the Experiments section).

\subsection{Reinforced Collaborative Distillation and Self-Learning}
RL agents are capable of discovering optimal policies through interaction with the environment, guided by carefully designed reward functions. This property makes RL well-suited for identifying and implementing more effective and adaptive training strategies for the student model. In our framework, the RL agent plays a crucial role, which can be summarized in two main aspects:

(1) Adaptive training difficulty: The agent modulates the difficulty of training samples to facilitate the student’s self-learning process. By progressively increasing the challenge level, the student is encouraged to handle more complex scenarios (see examples in Fig.~\ref{fig:sample}).

(2) Dynamic teacher guidance: During self-learning, the agent dynamically adjusts the degree of teacher guidance, enabling adaptive knowledge transfer based on the student’s current performance, allowing the student to more effectively unlock its potential.

In the following, we present a detailed description of the agent’s workflow. As shown in Fig. \ref{fig:overall}, the agent uses the student’s current performance and the teacher-student gap as input states. Based on these states, the agent produces actions to control the degradation level of the original images. Meanwhile, the agent also generates actions to adjust the strength of the teacher’s guidance (\textit{i.e.}, the values of $\alpha_t$ and $\alpha_s$ in Eq. \ref{eq:la}) to optimize the knowledge transfer.

After each episode of training, the teacher-student performance gap is re-evaluated and used as the reward for updating the agent’s parameters. This process is repeated until the student network achieves or exceeds the teacher’s performance. The definitions of state, action, and reward in the context of image fusion are detailed as follows.

\textbf{State.} 
In our framework, the state is defined based on both the performance of the student model and the gap between the teacher and student. Traditional methods typically rely on ground-truth labels to evaluate model performance. However, in image fusion, there is no physical ground-truth fused image, making it an inherently unsupervised task.

To address this challenge, we assess the performance of both the student and teacher models using a suite of quantitative metrics, including average gradient, edge intensity, visual information fidelity \cite{sheikh2006image}, standard deviation, and CLIP-IQA \cite{wang2023exploring}. These metrics are chosen to provide a comprehensive evaluation of image fusion quality from multiple perspectives.

The scores for each metric are first normalized to ensure comparability and then combined to form performance indicators for the student ($\mathbf{m}_s^k$) and teacher ($\mathbf{m}_t^k$) models at step $k$. The teacher-student gap is then defined as the difference between these indicators: $\Delta \mathbf{m}^k = \mathbf{m}_t^k - \mathbf{m}_s^k$. The overall state embedding at step $k$ is thus constructed as:
\begin{equation}
	\mathbf{s}^k = \left[ \mathbf{m}_s^k, \Delta \mathbf{m}^k \right].
\end{equation}
This formulation enables our method to represent both the absolute and relative performance of the student model in an unsupervised setting.

It is should be noted that standard CLIP-IQA evaluates image quality using generic prompts such as “\textit{good photo}” and “\textit{bad photo}”. However, these prompts do not capture important aspects specific to image fusion, such as object saliency and detail preservation. To better adapt CLIP-IQA for the IVIF task, we redesign its prompts to more accurately reflect fusion objectives. Specifically, we use “\textit{a clear image with detailed background and salient objects}” as the positive prompt and “\textit{an unclear image with minimal details and indistinct objects}” as the negative prompt. This enhanced version, denoted as IQA* in our paper, provides an evaluation that is more closely aligned with the goals of image fusion.

\begin{figure*}[t]
	\centering
	\includegraphics[width=\textwidth]{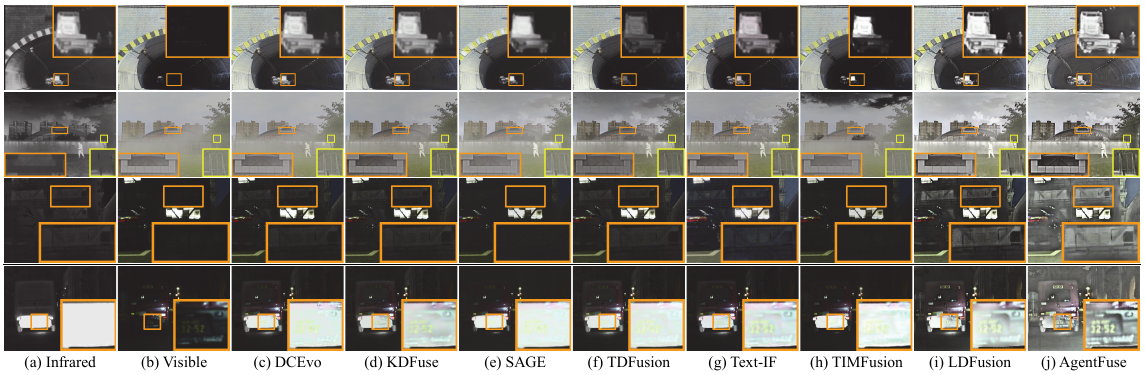} 
	\caption{Comparison of the fusion results from different methods. Text-IF uses its anti-degradation text for low light.}
	\label{fig:cp1}
\end{figure*}

\textbf{Action.} 
Conditioned on the state information, the agent outputs a continuous action vector that jointly determines the difficulty level of the training samples and the relative strengths of the teacher’s guidance and the student’s self-learning. Formally, the action vector is given by:
\begin{equation}
	\mathbf{a}^k = [\alpha_t^k, \alpha_s^k, \boldsymbol{\alpha}_d^k],
\end{equation}
where $a_t^k$ and $a_s^k$ (normalized via softmax) regulate the contributions of $\mathcal{L}_t$ (Eq. \ref{eq:lt}) and $\mathcal{L}_s$ (Eq. \ref{eq:ls}), respectively. $\boldsymbol{\alpha}_d^k$ modulates the difficulty of the training samples by controlling image degradation parameters. $\boldsymbol{\alpha}_d^k$ consists of five components, each controlling the kernel size for Gaussian blur, the degree of lossy compression, the range of color jitter (including brightness and contrast adjustments), and the level of added noise (see the examples in Fig. \ref{fig:sample}). 

In practice, the agent outputs the mean and variance parameters of a Gaussian distribution over the action space, from which the action vector is sampled. During training, the reparameterization trick is employed to enable backpropagation through the sampling operation.

\textbf{Reward.} 
The reward function is designed to evaluate how effectively each action improves the student model’s fusion performance relative to the teacher. In our framework, the reward not only encourages the student to close the performance gap with the teacher, but also motivates the student to eventually outperform the teacher.

Given that IVIF aims to integrate source images and enhance perceptual quality, we use two complementary metrics for assessment: visual information fidelity (VIF), which reflects information content, and the IQA* score, which reflects overall perceptual quality. Both metrics are normalized and then combined into a single evaluation score, denoted as $E$.

The reward at step $k$ is then calculated as the difference in $E$ between the student and teacher model outputs:
\begin{equation}
	r^k = E(\mathbf{I}_s^k) - E(\mathbf{I}_t^k),
\end{equation}
where $\mathbf{I}_s^k$ and $\mathbf{I}_t^k$ represent the current fused images produced by the student and teacher, respectively. Under this formulation, a positive reward indicates that the student has improved and even surpassed the teacher’s performance on the fused image, while a negative reward penalizes actions that increase the performance gap.

\textbf{Agent Optimization.} Since our framework involves continuous action spaces, we adopt policy-based reinforcement learning methods, which are more suitable than value-based approaches in this context. In our setting, the agent continuously adjust both the degree of teacher guidance and the difficulty level of training samples. Inspired by \cite{yang2025multi}, we employ a policy gradient algorithm, chosen for its simplicity and effectiveness in optimizing continuous decisions. Formally, at time step $k$, the agent observes the current state $s^k$ and produces a probability distribution over actions, denoted as $\pi_\Phi(a^k \mid s^k)$, where $\Phi$ represents the agent's parameters. The agent then samples an action from this distribution and receives a corresponding reward $r^k$. The parameters are updated to maximize the expected cumulative reward, using the standard policy gradient update rule:

\begin{equation}
	\label{eq:pg}
	\Phi \leftarrow \Phi - \alpha \mathbb{E}_{\pi_\Phi} \left[ \nabla_{\Phi} \log \pi_\Phi(a^k \mid s^k) R^k \right],
\end{equation}
where $\alpha$ is the learning rate, and $R^k$ denotes the cumulative reward over the next $p$ steps: ${R^k} = \sum_{i=0}^{p} r^{k+i}$ (we empirically find that $p=4$ provides a good balance between learning efficiency and stability). 

Our agent is a neural network with several linear layers with ReLU activations, followed by a linear layer that outputs the parameters of the action distribution. Through iterative optimization, the agent learns to output actions that enhance the student model's learning progress.

\begin{table*}[t]
	\centering
	\renewcommand{\arraystretch}{1.1}
	\setlength{\tabcolsep}{1.8pt}
	\begin{tabular}{l | c | c c c c c | c c c c c | c c c c c}
		\hline
		\multirow{2}*{Methods} & \multirow{2}*{Venue}
		& \multicolumn{5}{c|}{M3FD}
		& \multicolumn{5}{c|}{TNO}
		& \multicolumn{5}{c}{MSRS} \\
		\cline{3-17}
		& & AG $\uparrow$ & SF $\uparrow$ & VIFF $\uparrow$ & EI $\uparrow$ & EN $\uparrow$ & AG $\uparrow$ & SF $\uparrow$ & VIFF $\uparrow$ & EI $\uparrow$ & EN $\uparrow$ & AG $\uparrow$ & SF $\uparrow$ & VIFF $\uparrow$ & EI $\uparrow$ & EN $\uparrow$ \\
		\hline
		DCEvo     & \scalebox{0.9}{CVPR'25}    & 5.552 & 5.785 & 0.435 & 48.16 & 6.842 & 5.415 & 5.818 & 0.401 & 41.38 & 6.952 & 4.357 & 4.881 & 0.755 & 39.75 & 6.625 \\
		KDFuse    & \scalebox{0.9}{InfFus'25}      & 5.592 & 5.872 & 0.428 & 49.25 & 6.820 & 6.010 & 6.321 & 0.492 & 47.36 & 6.935 & 4.192 & 4.794 & 0.730 & 39.07 & 6.655 \\
		SAGE      & \scalebox{0.9}{CVPR'25}    & 5.079 & 5.541 & 0.488 & 44.14 & 6.850 & 4.769 & 5.339 & 0.509 & 37.46 & 7.043 & 3.659 & 4.375 & 0.590 & 33.27 & 6.002 \\
		TDFusion  & \scalebox{0.9}{CVPR'25}    & 5.815 & 5.926 & 0.557 & 50.65 & 6.991 & 6.382 & 6.643 & 0.724 & 48.26 & 7.224 & 4.352 & 4.890 & 0.805 & 39.73 & 6.734 \\
		Text-IF   & \scalebox{0.9}{CVPR'24}    & 6.077 & 6.072 & 0.491 & 52.91 & 6.902 & 6.035 & 6.288 & 0.542 & 48.15 & 7.222 & 4.404 & 4.914 & 0.774 & 40.48 & 6.643 \\
		TIMFusion & \scalebox{0.9}{TPAMI'24}    & 4.920 & 5.468 & 0.323 & 42.83 & 6.753 & 5.072 & 5.503 & 0.304 & 36.73 & 6.968 & 3.331 & 4.186 & 0.556 & 31.12 & 6.273 \\
		LDFusion  & \scalebox{0.9}{ACMMM'25}      & 8.297 & 7.029 & 0.709 & 65.40 & \textbf{7.342} & 7.772 & 7.262 & \textbf{0.868} & 59.64 & 7.423 & 6.547 & 6.380 & \textbf{1.137} & 60.75 & 7.259 \\
		\hline
		AgentFuse & Proposed   & \textbf{10.123} & \textbf{7.840} & \textbf{0.721} & \textbf{78.39} & 7.335
		& \textbf{8.769} & \textbf{7.922} & 0.755 & \textbf{67.57} & \textbf{7.469}
		& \textbf{7.676} & \textbf{6.935} & 1.039 & \textbf{61.03} & \textbf{7.367} \\
		\hline
	\end{tabular}
	\caption{Quantitative assessment of our AgentFuse and compared methods on M3FD, TNO and MSRS (Bold: the best results).}
	\label{tab:cp1}
\end{table*}

\begin{table}[t]
	\centering
	\renewcommand{\arraystretch}{1.1}
	\begin{tabular}{lccc}
		\hline
		Methods     & mAP@0.5 & mAP@0.75 & mAP@0.5:0.95 \\
		\hline
		DCEvo       & 0.784  & 0.547  & 0.520 \\
		KDFuse      & 0.781  & 0.554  & 0.524 \\
		SAGE        & 0.784  & 0.548  & 0.525 \\
		TDFusion    & 0.783  & 0.553  & 0.525 \\
		Text-IF     & 0.782  & 0.549  & 0.523 \\
		TIMFusion   & 0.779  & 0.546  & 0.519 \\
		LDFusion    & \textbf{0.794}  & 0.549  & 0.529 \\
		\hline
		AgentFuse   & 0.791  & \textbf{0.562} & \textbf{0.532}  \\
		\hline
	\end{tabular}
	\caption{Quantitative comparison of object detection performance on M3FD (Bold: the best results).}
	\label{tab:detect}
\end{table}

\begin{table*}[t]
	\centering
	\renewcommand{\arraystretch}{1.1}
	\setlength{\tabcolsep}{7pt}
	\begin{tabular}{lcccccccc}
		\hline
		& DCEvo & KDFuse & SAGE & TDFusion & Text-IF & TIMFusion & LDFusion & AgentFuse \\
		\hline
		Params (M)     & 2.97     & 2.49     & 0.13     & \textbf{0.059}   & 215.11  & 1.23    & 5.78  & 0.65   \\
		FLOPs  (G)     & 149.05   & 125.02   & 3.32     & 2.97    & 249.71  & 35.26   & 13.09 & \textbf{2.71}   \\
		M3FD   (ms)    & 2427.27  & 6213.33  & 32.86    & 180.560 & 1748.92 & 218.28  & 52.87 & \textbf{17.61}  \\
		TNO    (ms)    & 919.37   & 1057.70  & 14.87    & 69.54   & 960.41  & 85.52   & 32.69 & \textbf{11.26}  \\
		MSRS   (ms)    & 993.90   & 525.99   & 13.91    & 68.88   & 763.00  & 43.98   & 20.83 & \textbf{12.21}   \\
		\hline
	\end{tabular}
	\caption{Efficiency comparison of different fusion models on M3FD, TNO and MSRS (Bold: the best results). }
	\label{tab:speed}
\end{table*}

\subsection{Experiments}
\subsection{Experimental Settings}
\textbf{Implementation Details.}
The model is implemented using PyTorch on an Intel(R) Xeon(R) 4214 CPU and Nvidia 3090 GPUs. For training the student model, we use the Adam optimizer with a learning rate of 0.002 and a batch size of 24. The agent learning rate $\alpha$ in Eq. \ref{eq:pg} is set to 0.01. For teacher-guidance loss $\mathcal{L}_t$, we use the feature maps from layers ``conv1\_1'', ``conv2\_1'', ``conv3\_1'', ``conv4\_2'', and ``conv5\_2'' as the set of $m$ to compute this loss. The student model is pre-trained for 20 epochs using only $\mathcal{L}_t$, and subsequently trained for 100 epochs with total loss $\mathcal{L}_a$. 

\textbf{Datasets.}
We employ three publicly available datasets in our experiments: M3FD \cite{liu2022target}, TNO \cite{toet2017tno} and MSRS\cite{tang2022piafusion}. The network is trained on 2,940 image pairs from the M3FD dataset, with an additional 300 pairs from M3FD used for testing fusion performance. For further evaluation, we use the standard test split from the MSRS dataset. Additionally, commonly used image pairs from the TNO dataset are adopted to assess the generalization of our method. For color visible images, we convert them into the YCbCr color space, fuse the brightness component with the infrared image, and then restore the chrominance components to the fused image.

\textbf{Metrics.}
The evaluation metrics include average gradient (AG) \cite{cui2015detail}, spatial frequency (SF) \cite{roberts2008assessment}, visual information fidelity for fusion (VIFF) \cite{han2013new}, edge intensity (EI) \cite{zhang2025ddbfusion}, and information entropy (EN)\cite{roberts2008assessment}. Among them, AG reflects degree of detail richness. SF quantifies overall clarity. VIFF measures the perceptual quality. EI reflects the preservation of structural information. EN quantifies the amount of information contained in the fused result. In general, a higher value of these metrics indicates better fusion performance. 

\textbf{Competitors.} 
Seven state-of-the-art methods are used to conduct qualitative and quantitative comparisons, including DCEvo \cite{liu2025dcevo}, KDFuse \cite{yang2025kdfuse}, SAGE \cite{wu2025every}, TDFusion \cite{bai2025task}, Text-IF \cite{yi2024text}, TIMFusion \cite{liu2024task} and LDFusion \cite{wang2024infrared}.

\subsection{Comparison with State-of-the-art Methods}
\label{sec:compare}
\textbf{Qualitative Comparison.}
Some qualitative results of our AgentFuse and seven state-of-the-art (SOTA) methods are presented in Fig. \ref{fig:cp1}. As illustrated by the example in the first row, it can be seen that our method significantly enhances both object saliency and detail clarity. Although DECvo, KDFuse, and LDFusion preserve key details from the source images, it is evident that our approach generates fusion results that are clearer and richer in detail.

Regarding object-level fusion performance, while DCEvo, SAGE, and TIMFusion leverage high-level tasks (such as object detection and semantic segmentation) to supervise the fusion process, they still do not achieve object fusion quality comparable to ours (see the pedestrians highlighted in orange boxes). Compared to the teacher model (LDFusion), our method produces fusion results with higher information fidelity (see the truck in the orange box) and more salient targets (see the pedestrians).

\textbf{Comparison on Challenging Scenes.}
We evaluate the performance of our method in challenging scenarios, including foggy and low-light conditions.

In foggy conditions, our AgentFuse can more effectively integrate information from the source images, producing clearer fused images with sharp textures and edges, while also excluding irrelevant information such as fog. As shown in the second example in Fig. \ref{fig:cp1}, the structure of the building (orange box) and the ladder (yellow box) are significantly enhanced in our fusion results. In contrast, the other methods produce suboptimal results with relatively lower clarity.

We also assess fusion performance under low-light conditions. In the last two rows of Fig. \ref{fig:cp1}, our method achieves much better fusion results compared to the other methods. In particular, a close-up view of the building (orange box) shows that our approach successfully fuses and enhances the details and structures from the source images, enabling the fused images to reveal visual information that is weak or nearly invisible in the original images. This results in a vivid appearance and demonstrates the impressive power of collaborative distillation and self-learning. The other methods produce relatively inferior results, although Text-IF was specifically designed with degradation-aware strategies.

\textbf{Quantitative Comparison.}
Table \ref{tab:cp1} summarizes the quantitative results of our AgentFuse and the compared state-of-the-art (SOTA) methods across various evaluation metrics. Our method achieves the highest average scores on almost all metrics across the M3FD, TNO, and MSRS datasets. Our method significantly outperforms the compared methods on metrics such as AG, SF, and EI. These improvements demonstrate that our method can integrate more details and structures from the source images, leading to more informative fusion results. Moreover, the higher VIFF obtained by our method reflects superior information fidelity and improved perceptual quality in the fusion results. The consistent and high performance across multiple datasets demonstrates the strong generalization capability of our method.

\begin{figure*}[t]
	\centering
	\includegraphics[width=18cm]{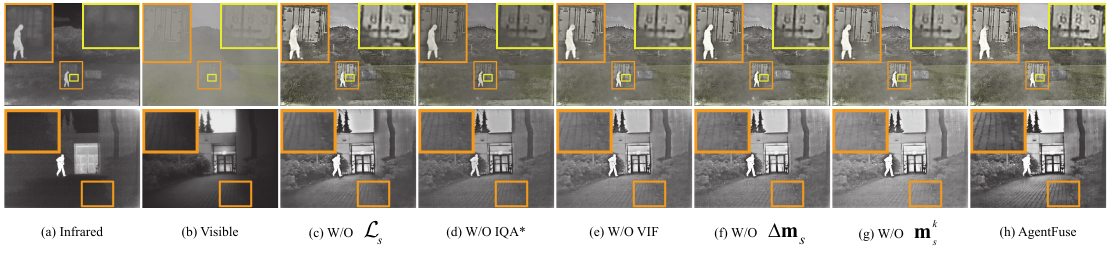} 
	\caption{Some examples of the fusion results from the ablation experiments.}
	\label{fig:ab1}
\end{figure*}

\textbf{Performance on High-level Task.}
To further evaluate our method, we conduct object detection experiments using the fused images generated by different methods. We employ YOLOv8 as the object detection baseline, training separate YOLOv8 models on the fusion results from each fusion method. The quantitative detection results are presented in Table \ref{tab:detect}. Our method achieves the better object detection performance with the highest scores in almost all metrics, despite obtaining a slightly lower but comparable mAP@0.5 score to LDFusion. These results indicate that our method can integrate more semantic information from the source images. Although DCEvo, SAGE, and TIMFusion leverage high-level tasks such as object detection or utilize large models like the Segment Anything Model to supervise the fusion process, their detection performance is still inferior to ours.

\textbf{Efficiency Analysis.}
Table \ref{tab:speed} summarizes the parameters, computational overhead, and inference speed across different datasets. Compared with the teacher model (LDFusion), the number of parameters in our method is only 11.2\% of the teacher’s, and the computational overhead is just 20.7\% of the teacher’s. Even so, our method achieves much better fusion quality compared to SOTA competitors, including the teacher model. Our approach also significantly improves inference speed, achieving real-time performance even on a notebook equipped with an RTX 3070 laptop GPU (see Table~\ref{tab:speed}). Thanks to the simple architecture and low computational cost of our AgentFuse achieves substantially faster inference speeds than existing methods, enabling real-time performance even on high-resolution inputs (\textit{e.g.}, M3FD). This highlights the practicality and deployment potential of our approach in real-world applications.

Although TDFusion has the smallest number of parameters among all compared models, its inference speed is significantly slower than that of our AgentFuse. This may be because TDFusion stacks many layers in a strictly serial manner, introducing strong inter-layer dependencies and limiting opportunities for parallel execution on GPUs.

\subsection{Ablation Study}
\begin{table}[t]
	\centering
	\renewcommand{\arraystretch}{1.1}
	\begin{tabular}{lccccc}
		\hline
		& AG $\uparrow$    & SF $\uparrow$   & VIFF $\uparrow$ & EI $\uparrow$    & EN $\uparrow$   \\
		\hline
		W/ O $\mathcal{L}_s$         & 6.776 & 6.611 & 0.752 & 54.071 & 7.013 \\
		W/ O IQA*                    & 7.248 & 6.528 & 0.674 & 57.333 & 6.806 \\
		W/ O VIF                     & 7.759 & 6.748 & 0.739 & 60.192 & 6.895 \\
		W/ O $\Delta \mathbf{m}^k$   & 8.359 & 7.095 & 0.835 & 66.450 & 7.148 \\
		W/ O $\mathbf{m}_s^k$          & 7.720 & 6.863 & 0.776 & 60.806 & 7.038 \\
		\hline
		AgentFuse                    & \textbf{8.785} & \textbf{7.391} & \textbf{0.883} & \textbf{68.807} & \textbf{7.361} \\
		\hline
	\end{tabular}
	\caption{Summary of ablation experiment results on M3FD, TNO and MSRS (Bold: the best results).}
	\label{tab:ab}
\end{table}

To evaluate the effectiveness of the proposed methods, we perform ablation studies to evaluate the effectiveness of various components within the proposed framework using the TNO, M3FD, and MSRS datasets. Table \ref{tab:ab} summarizes the results of these ablation experiments, while Fig. \ref{fig:ab1} provides some examples for visual comparison. 

The results demonstrate the crucial role of self-learning in improving image fusion quality. Specifically, when the self-learning loss ($\mathcal{L}_s$) is removed, there is a noticeable drop in performance across all metrics, particularly AG, SF, and EI, indicating that self-learning significantly enhances image clarity and detail sharpness (see the building in second row of Fig. \ref{fig:ab1}(c)). Removing IQA* from the reward function also leads to inferior fusion results with low clarity (see the yellow boxes of Fig. \ref{fig:ab1}(d)), highlighting its importance in guiding the model to optimize perceptual quality. Furthermore, the absence of VIF in the reward function results in a substantial decrease in SF and EI, emphasizing its role in preserving source information, such as texture and edges. When the student state ($\mathbf{m}_s^k$) or teacher-student gap ($\Delta \mathbf{m}^k$) is removed from the agent’s input, the performance also declines, suggesting that these components are vital for the agent’s ability to differentiate between the teacher’s guidance and the student’s self-learning progress.

\subsection{Conclusion}
In our paper, we propose a novel collaborative distillation and self-learning framework for image fusion, which drives the student’s self-learning via an RL agent. The agent adaptively modulates the difficulty of the training samples by degrading image quality, encouraging the student to develop more robust fusion strategies under challenging conditions. Furthermore, during self-learning, the agent dynamically adjusts the strength of the teacher’s guidance to promote adaptive knowledge transfer. Through this dynamic interplay, the student model can unlock its potential more effectively. Extensive experiments show that our framework significantly improves the performance and generalization ability of image fusion models, highlighting the potential of reinforcement learning and self-learning in advancing IVIF performance.

\clearpage
\bibliography{fusion}

\clearpage


\end{document}